\documentclass[11pt,letterpaper]{article}

\usepackage[margin=1in]{geometry}
\usepackage{amssymb}

\usepackage{natbib}
\usepackage{wrapfig}
\usepackage{graphicx}
\usepackage{booktabs}
\usepackage[table]{xcolor} 
\usepackage[T1]{fontenc}
\usepackage{multirow}
\usepackage[most]{tcolorbox}
\usepackage{authblk} 

\tcbset{
    colback=gray!5,
    colframe=gray!60,
    boxrule=0.3pt,
    arc=2pt,
    left=6pt,
    right=6pt,
    top=6pt,
    bottom=6pt
}

\usepackage{tikz}
\definecolor{lightblue}{rgb}{0.93,0.95,1.0} 
\definecolor{perceivercolor}{RGB}{108, 142, 191}
\definecolor{vtccolor}{RGB}{218, 116, 105} 
\definecolor{perceiverHighlight}{RGB}{240, 250, 255}
\definecolor{dtwcolor}{RGB}{27, 183, 252}
\definecolor{mamcolor}{RGB}{255, 142, 5}

\usepackage{hyperref}
\hypersetup{
    colorlinks=true,
    linkcolor=blue,
    filecolor=magenta,      
    urlcolor=cyan,
    citecolor=blue,
}

\usepackage{orcidlink}

\begin{document}

\title{InstrAction: Action-Centric Pretraining for Instructional Videos} 

\author[1]{Zhuoyi Yang}
\author[1]{Jiapeng Yu}
\author[2]{Reuben Tan}
\author[3]{Boyang Li}
\author[1]{Huijuan Xu}

\affil[1]{Pennsylvania State University, University Park, USA}
\affil[2]{Microsoft Research}
\affil[3]{Nanyang Technological University, Singapore}

\date{} 
\maketitle

\begin{figure}[!ht] 
  \centering 
  \includegraphics[width=0.7\columnwidth]{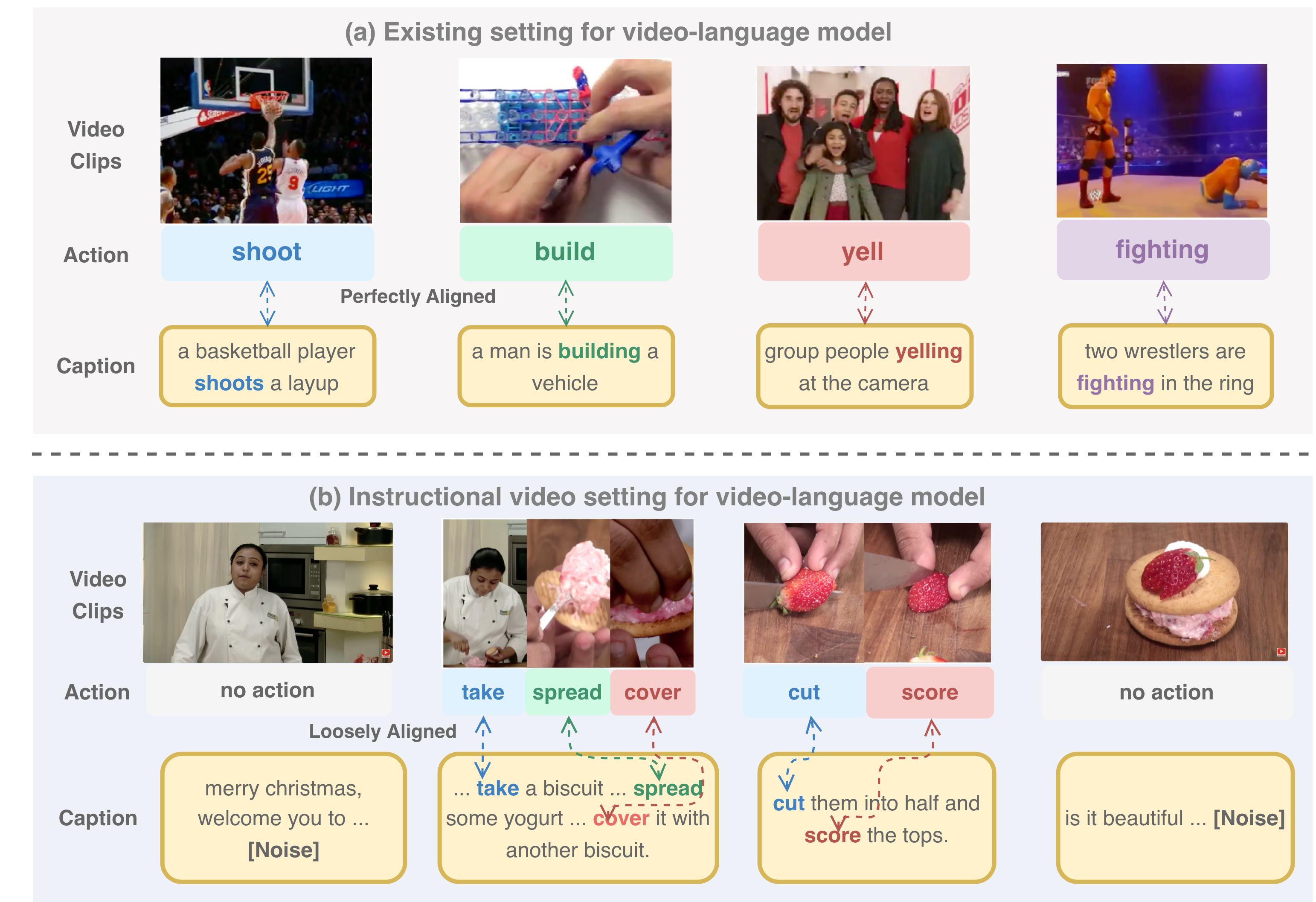}
  \caption{\textbf{Video-language modeling paradigms.} (a) Existing paradigms use trimmed clips with single atomic actions and perfectly aligned captions. (b) Instructional video setting involves sequential actions, loosely aligned descriptions, and non-instructional noise.}
  \label{fig:experiments}
  \vspace{-1em}
\end{figure}

\begin{abstract}
Understanding instructional videos requires recognizing fine-grained actions and modeling their temporal relations, which remains challenging for current Video Foundation Models (VFMs). This difficulty stems from noisy web supervision and a pervasive “static bias”, where models rely on objects rather than motion cues. To address this, we propose \textbf{InstrAction}, a pretraining framework for instructional videos' action-centric representations. We first introduce a data-driven strategy, which filters noisy captions and generates action-centric hard negatives to disentangle actions from objects during contrastive learning. At the visual feature level, an Action Perceiver extracts motion-relevant tokens from redundant video encodings. Beyond contrastive learning, we introduce two auxiliary objectives: Dynamic Time Warping alignment (DTW-Align) for modeling sequential temporal structure, and Masked Action Modeling (MAM) for strengthening cross-modal grounding. Finally, we introduce the InstrAct Bench to evaluate action-centric understanding, where our method consistently outperforms state-of-the-art VFMs on semantic reasoning, procedural logic, and fine-grained retrieval tasks. \textbf{Code will be released upon acceptance.}

\vspace{0.5em}
\end{abstract}

\section{Introduction}
\label{sec:intro}

Early research in video understanding primarily focused on trimmed clips, typically formulated as supervised action recognition on standard benchmarks~\cite{schuldt2004recognizing,soomro2012ucf101,kuehne2011hmdb,kay2017kinetics}. With the emergence of large-scale instructional video datasets~\cite{miech19howto100m,tang2019coin} and the success of cross-modal contrastive learning in image–text models~\cite{radford2021learning}, research has increasingly shifted toward video–language pretraining using long, untrimmed instructional videos~\cite{miech19endtoend,bain2021frozen,xu2021videobert,yu2022videoclip}. Unlike trimmed clips with a single action, instructional videos depict procedures composed of sequential actions. Consequently, understanding instructional videos requires not only recognizing individual actions, but also modeling their temporal organization as coherent procedures.

However, recent studies show that current video–language models exhibit limited understanding of verbs and actions. Instead, these models rely heavily on static visual cues (e.g., objects and backgrounds) and textual cues (e.g., nouns), failing to capture true action semantics~\cite{hendricks2021probing,park2022exposing,yuksekgonul2023when}. To mitigate this issue, prior work has explored strategies such as generating action-focused hard negatives~\cite{wang2023paxion} and designing verb-aware contrastive objectives~\cite{momeni2023verbs}.

Nevertheless, these approaches are primarily developed for trimmed videos~\cite{monfort2021smit,goyal2017something,grauman2022ego4d} that contain single or simultaneous actions and are paired with clean captions. In contrast, instructional videos rely on ASR transcripts that describe continuous sequences of actions. Such transcripts are often noisy and loosely aligned with visual content~\cite{miech19howto100m,tang2019coin}. As a result, the supervision signals derived from these transcripts are substantially weaker, making it difficult to disentangle multiple actions and ground the action semantics in videos. These challenges indicate that simply extending trimmed videos' techniques is insufficient for instructional video understanding. To extend action-centric modeling beyond the trimmed-video paradigm, we propose \textbf{InstrAction}, a pretraining framework for instructional videos' action-centric representations. Our framework combines a data-driven strategy with an action-centric model design. 

On the data side, to address the noisy supervision of ASR transcripts, we develop an LLM-assisted curation pipeline that filters non-instructional captions and extracts verb phrases for fine-grained temporal supervision. We further construct action-centric hard negatives (HNs) to disentangle actions from static objects: verb-altered HNs penalize incorrect action identification, while order-swapped HNs enforce procedural temporal reasoning. This curated data provides strong action-centric training signals for contrastive learning.

On the model side, to mitigate the static bias of video encoders, we introduce an Action Perceiver network to distill action related representations from redundant video features. The module compresses dense video embeddings into a compact set of latent action tokens using a Perceiver-based resampler~\cite{jaegle2021perceiver,alayrac2022flamingo}. To ensure that these tokens capture true action semantics, we further introduce a verb-guided distillation mechanism~\cite{li2021albef}. Verb phrase embeddings attend to video features to produce teacher tokens that capture action semantics, while the Perceiver's output latents act as students. This guidance encourages the latent space to focus on action-relevant cues, enabling robust action representations even when verb phrases are unavailable during downstream inference.

Beyond global contrastive video–text alignment, we introduce two auxiliary objectives to better model the temporal and cross-modal structure of instructional videos. To capture sequential action structure, we propose DTW-Align, a Dynamic Time Warping based objective that models temporal alignment between action tokens and extracted verb phrases, enabling flexible, non-linear correspondence between visual action dynamics and textual descriptions~\cite{cuturi2017soft}. To further strengthen cross-modal grounding under noisy transcripts, we introduce Masked Action Modeling (MAM), inspired by masked language modeling~\cite{devlin2019bert}. By masking action tokens in captions and predicting them using both visual evidence and textual context, MAM encourages deeper integration between visual motion cues and language representations.

\vspace{0.5em}
\noindent\textbf{Contributions.} In summary, we propose InstrAction, a comprehensive pretraining framework for instructional videos that integrates action-centric data curation, a action-distilling module, and dedicated self-supervised objectives:

\begin{itemize}
    \item \textbf{Data-Driven Training \& Curation:} We develop an LLM-assisted data curation pipeline to filter non-instructional noise and integrate action-centric hard negatives. Based on this curated data, we also introduce the InstrAct Bench to diagnose action-centric understanding.
    \item \textbf{Motion-Distilling Architecture:} We propose the Action Perceiver to aggregate motion-essential cues from redundant video embeddings, further refined by a verb-guided distillation mechanism to ensure the latent space is action-centric.
    \item \textbf{Multi-Action Learning Objectives:} We introduce DTW-Align for non-linear action-verb alignment and MAM to strengthen cross-modal action grounding.
\end{itemize}

\section{Related Work}

\subsection{Action Understanding in Vision–Language Models} 

Static cues like background scenery and object appearances are well-known shortcut features in action and video understanding~\cite{li2023stillmix,lee2025dancce,fioresi2025albar}, and vision-language models are not immune to them. Hendricks and Nematzadeh~\cite{hendricks2021probing} found that image-text models tend to understand verbs using object occurrence rather than relational and temporal cues. In the video domain, Park et al.~\cite{park2022exposing}showed that even with multiple frames, video-text models rely heavily on static cues rather than action dynamics and interactions. 

To mitigate this issue, action-sensitive datasets are built to provide strong supervision on actions. The Action Dynamics Benchmark~\cite{wang2023paxion} created hard negatives by reversing individual actions and entire video clips. VFC~\cite{momeni2023verbs} generate negative captions with a different verb for each video clip. These methods are designed for short videos that contain single atomic action. In contrast, instructional videos consist of diverse and compositional actions with complex causal and temporal dependencies. To our best knowledge, this is the first paper tackling the static shortcut issue in instructional video understanding. 

\subsection{Instructional Video Datasets}

Instructional videos feature multi-step, compositional activities with strong temporal dependencies. Early curated datasets such as YouCook2 \cite{zhou2018towards} and CrossTask \cite{zhu2019cross} provide high-quality annotations for procedure segmentation and step localization. Others like EPIC-KITCHENS \cite{damen2018epic} and ProceL \cite{ehsan2019procel} focus on fine-grained actions and hand-object interactions. To improve generalization, larger corpora like COIN \cite{tang2019coin} and HowTo100M \cite{miech19howto100m} offer diverse domains for large-scale pretraining. However, their automated transcripts often contain noise or off-topic content, leading to weak semantic alignment between text and visual actions.

\section{Data Curation}
\label{sec:data}

\begin{figure}[!htbp]
    \centering
    \includegraphics[width=\textwidth]{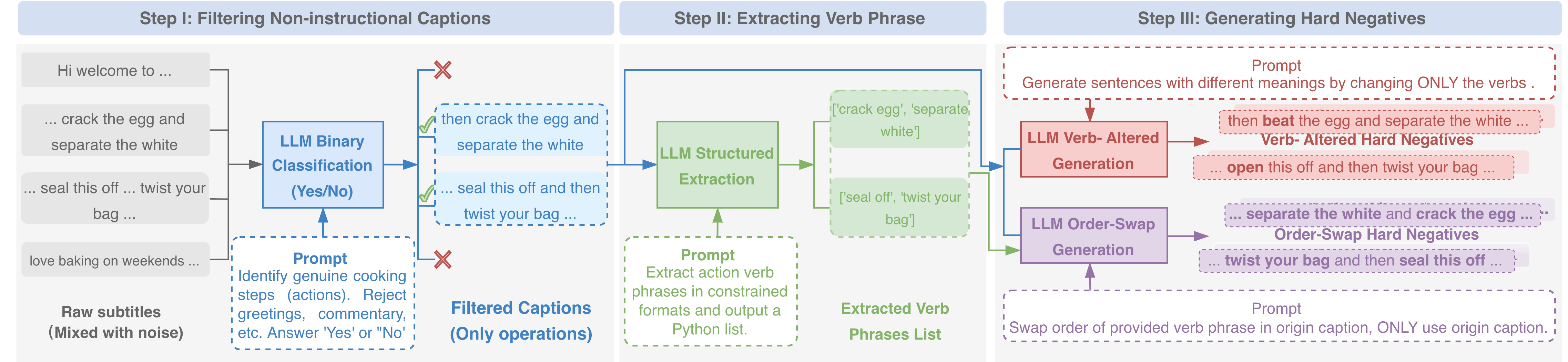}
    \caption{\textbf{Overview of the LLM-assisted data curation pipeline.} Our data-driven strategy consists of three stages: (I) filtering non-instructional noise from raw subtitles; (II) parsing the filtered captions to extract discrete action units into predefined structured formats; and (III) generating action-centric hard negatives. These include verb-altered negatives and order-swapped negatives.}
    \label{fig:framework}
\end{figure}

To construct action-centric supervision for instructional video pretraining, we develop an LLM-assisted data curation pipeline to refine the cooking subset of HowTo100M~\cite{miech19howto100m}. As illustrated in Figure~\ref{fig:framework}, the pipeline consists of three stages: (1) filtering non-instructional captions, (2) extracting structured verb phrases, and (3) generating action-centric hard negatives. Detailed prompts and implementation details are provided in the Appendix.

\paragraph{Filtering Non-Instructional Captions}
Raw subtitles in HowTo100M contain conversational filler and other non-instructional content that lacks visual grounding. To remove such noise, we employ an LLM-based binary classifier that distinguishes instructional steps from conversational text. Only captions describing concrete food-related actions are retained.

\paragraph{Verb Phrase Extraction}
To obtain structured supervision for temporal action modeling, we employ an LLM to parse captions and extract discrete verb phrases. We design prompts that structure actions into predefined syntactic patterns: ``V + N'', ``V-ing'', or ``V + Prep + N''. 

For example, a caption such as \textit{"Crack the egg using the side of this ceramic mug and separate the white into the small container."} is converted into a standardized action list: \textit{['crack egg', 'separate white']}. This structured decomposition isolates motion cues from redundant objects and provides supervision anchors for temporal alignment and hard negative generation.

\paragraph{Generating Action-Centric Hard Negatives}
To discourage reliance on static visual cues, we generate verb-altered negatives by replacing the main action verb with a semantically different one while keeping the rest of the caption unchanged. This forces the model to distinguish actions based on motion rather than static context.

To enforce procedural temporal reasoning, we further generate order-swapped negatives by permuting the chronological order of actions while preserving the original vocabulary (e.g., transforming \textit{"crack the egg and then whisk it"} into \textit{"whisk it and then crack the egg"}). This penalizes bag-of-words shortcuts and encourages the model to capture sequential action dependencies.

\section{Action Centric Model}

\begin{figure}[!htbp]
    \centering
    \includegraphics[width=0.95\textwidth]{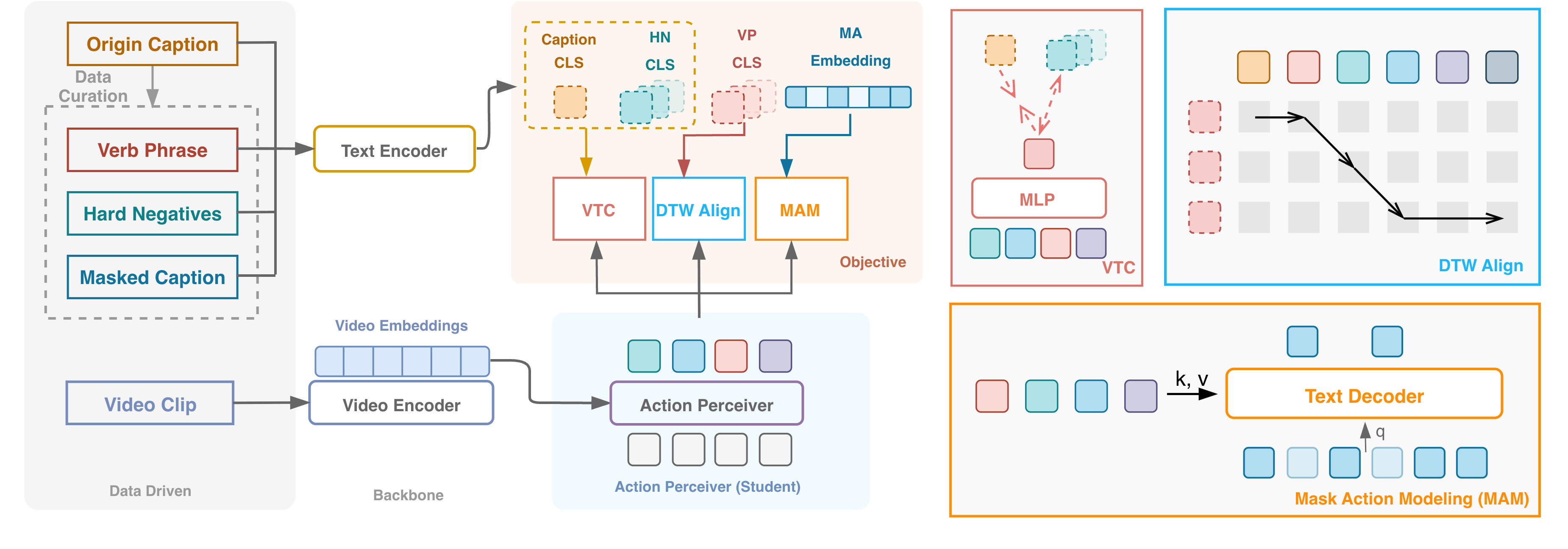}
    \caption{\textbf{Overview of the InstrAction framework.} Our model builds on a video–text backbone and an Action Perceiver that extracts action-centric representations. The model is trained with three objectives: video–text contrastive learning, DTW-Align for temporal grounding, and Masked Action Modeling (MAM) for cross-modal action understanding.}
    \label{fig:acm_framework}
\end{figure}

We next introduce the action-centric model architecture of InstrAction. The model builds on a video–text backbone to extract initial cross-modal representations, which are further refined by specialized modules for action understanding.

Specifically, we introduce an Action Perceiver network with verb-guided distillation (Sec.~\ref{sec:perceiver}) to extract action-relevant cues from redundant video features. To model the temporal dependencies of sequential actions, we incorporate a DTW-based Frame–Action Alignment objective (Sec.~\ref{sec:dtw}). To further strengthen cross-modal grounding, we adopt a Masked Action Modeling objective (Sec.~\ref{sec:mam}) that requires the model to recover masked action tokens conditioned on visual evidence.

\subsection{Action Perceiver with Knowledge Distillation}
\label{sec:perceiver}

\begin{wrapfigure}[14]{r}{0.5\textwidth}
  \centering
  \vspace{-20pt} 
  \includegraphics[width=\linewidth]{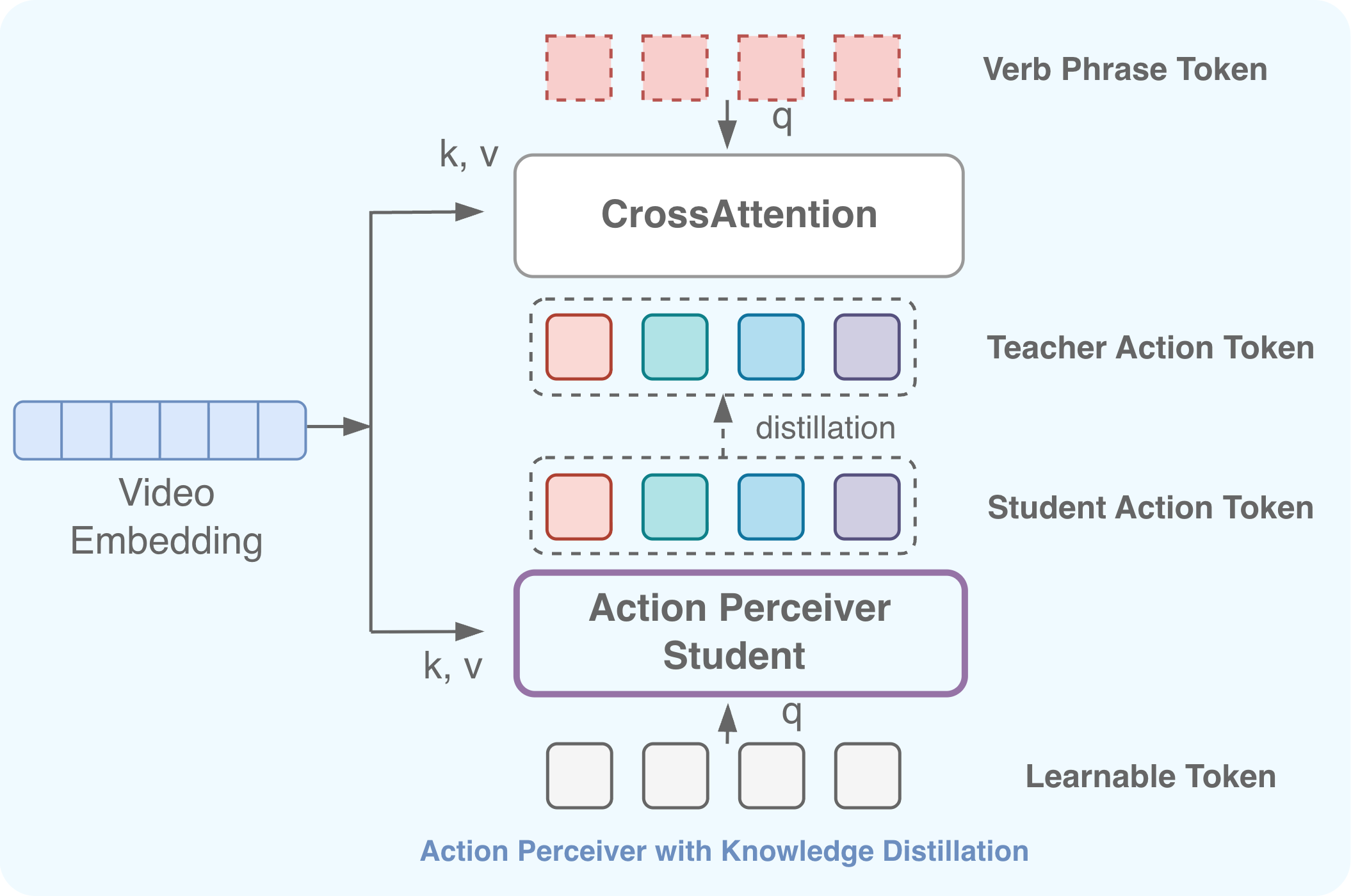}
  \caption{Action Perceiver module.}
  \label{fig:perceiver}
\end{wrapfigure}

To mitigate the static bias of video encoders in instructional videos, we introduce an Action Perceiver network optimized via verb-guided distillation. The module compresses redundant visual features into compact motion-aware Action Tokens.

\paragraph{Action Token Extraction}

We employ a Perceiver~\cite{jaegle2021perceiver} network to compress the visual features into a compact set of latent action representations. Formally, the visual embeddings are denoted as $\mathbf{V} \in \mathbb{R}^{T \times N \times C}$, where $T$, $N$, and $C$ denote the number of frames, patches per frame, and feature dimension. A set of $K$ learnable latent queries $\mathbf{L} \in \mathbb{R}^{K \times C}$ interacts with the flattened visual features via cross-attention to produce Student Action Tokens $\mathbf{S} \in \mathbb{R}^{K \times C}$:

\begin{equation}
\mathbf{S} = \text{Perceiver}(q=\mathbf{L}, k=\mathbf{V}, v=\mathbf{V}).
\end{equation}

To preserve temporal structure, we ensure each latent query aggregates localized motion dynamics by restricting each latent query to attend to a local temporal window rather than the entire sequence. This is implemented by injecting temporal embeddings and applying an attention mask to the similarity matrix.

\paragraph{Verb-guided Distillation}

To guide the latent tokens toward action semantics, we introduce a verb-guided distillation stream acting as a teacher. As illustrated in Figure~\ref{fig:perceiver}, extracted verb phrase embeddings $\mathbf{P} \in \mathbb{R}^{M \times C}$ attend to visual features to produce Teacher Action Tokens $\mathbf{T}$:

\begin{equation}
\mathbf{T} = \text{Perceiver}(q=\mathbf{P}, k=\mathbf{V}, v=\mathbf{V}).
\end{equation}

Since $\mathbf{T}$ is conditioned on explicit action verbs, it provides a semantic reference for motion. The student tokens $\mathbf{S}$ are then optimized to match the teacher through a soft contrastive distillation objective.

Let $\mathbf{s}$ denote similarity scores predicted from the student tokens and $\mathbf{s}'$ the teacher targets. The unidirectional distillation loss is defined as:

\begin{equation}
\mathcal{L}_{\text{single}}(\mathbf{s}, \mathbf{s}') =
-(1-\alpha) \sum \mathbf{y} \log(\sigma(\mathbf{s}))
-\alpha \sum \sigma(\mathbf{s}') \log(\sigma(\mathbf{s})),
\end{equation}

where $\sigma(\cdot)$ is the Softmax function and $\alpha$ controls the distillation intensity. The final bidirectional objective sums the losses over both video-to-text ($v2t$) and text-to-video ($t2v$) directions:
\begin{equation}
\mathcal{L}_{distill} = \mathcal{L}_{\text{single}}(\mathbf{s}_{v2t}, \mathbf{s}'_{v2t}) + \mathcal{L}_{\text{single}}(\mathbf{s}_{t2v}, \mathbf{s}'_{t2v}).
\end{equation}
This distillation guides the latent tokens toward verb-centric motion cues, enabling robust action representations even when verb phrases are unavailable during inference.

\subsection{DTW-based Action--Verb Alignment}
\label{sec:dtw}

To model temporal correspondence between visual actions and verb phrases, we design a DTW-based alignment objective that aligns the action token sequence with the extracted verb sequence.

\paragraph{Alignment Formulation}

Let $\mathbf{S} \in \mathbb{R}^{K \times C}$ denote the sequence of Action Tokens extracted by the Action Perceiver, forming a temporally ordered representation of video dynamics. The curated verb phrases are encoded to a sequence of embeddings $\mathcal{V} = \{\mathbf{v}_1, \dots, \mathbf{v}_M\}$, where $\mathbf{v}_j \in \mathbb{R}^C$.

We compute the alignment cost between $\mathbf{s}_i$ and $\mathbf{v}_j$ using cosine distance

\[
d(\mathbf{s}_i, \mathbf{v}_j) = 1 - \frac{\mathbf{s}_i \cdot \mathbf{v}_j}{\|\mathbf{s}_i\| \|\mathbf{v}_j\|}
\]

which yields a cost matrix $\mathbf{C} \in \mathbb{R}^{K \times M}$. The goal of DTW is to find the monotonic path through $\mathbf{C}$ with minimal cumulative cost. To enable end-to-end optimization, we adopt the differentiable Soft-DTW~\cite{cuturi2017soft} and define the alignment loss as

\begin{equation}
\mathcal{L}_{align}(\mathbf{S}, \mathcal{V}) = \text{Soft-DTW}(\mathbf{C}).
\end{equation}

\paragraph{Order-Aware Regularization}

Optimizing $\mathcal{L}_{align}$ alone may lead to degenerate alignments. To enforce temporal sensitivity, we introduce a contrastive regularization term inspired by~\cite{xue2023learning}. The intuition is that aligning the correct action order should yield lower cost than aligning a temporally disrupted sequence.

We construct a negative action sequence $\tilde{\mathbf{S}}$ by reversing the token order (i.e., $\tilde{\mathbf{S}} = [\mathbf{s}_K, \dots, \mathbf{s}_1]$). A hinge loss enforces a margin between the positive and negative alignments:

\begin{equation}
\mathcal{L}_{order} =
\max\left(
0,
\mathcal{L}_{align}(\mathbf{S}, \mathcal{V})
-
\mathcal{L}_{align}(\tilde{\mathbf{S}}, \mathcal{V})
+
\beta
\right).
\end{equation}
where $\beta$ is a margin hyperparameter. This objective explicitly penalizes the model if it fails to distinguish the correct procedural flow from a reversed sequence.

\paragraph{Total Objective}

The final DTW objective is computed over the batch:

\begin{equation}
\mathcal{L}_{dtw} =
\frac{1}{B}
\sum_{i=1}^{B}
\left(
\mathcal{L}_{align}^{(i)}
+
\gamma \mathcal{L}_{order}^{(i)}
\right).
\end{equation}

\subsection{Masked Action Modeling}
\label{sec:mam}
While DTW-Align captures temporal correspondence between actions and verbs, it operates under a dual-encoder contrastive paradigm and provides limited token-level cross-modal interaction. To address this limitation, we introduce Masked Action Modeling (MAM), which requires the model to reconstruct masked action verbs by attending to visual Action Tokens $\mathbf{S}$.

\paragraph{Multimodal Decoding}
We employ a multimodal text decoder~\cite{yu2022coca} to predict masked caption tokens. For each layer $l$, textual hidden states $\mathbf{h}_t^{(l)}$ are first processed by causal self-attention to capture linguistic context, followed by cross-attention that integrates visual Action Tokens $\mathbf{S}$:
\begin{equation}
\mathbf{h}_t^{(l)} = \text{Cross-Attn}(\text{Self-Attn}(\mathbf{h}_t^{(l-1)}), \mathbf{S})
\end{equation}

\paragraph{Learning Objective}
The decoder is optimized using an auto-regressive objective. Let $\mathbf{h}_t$ denote the textual hidden state at step $t$. The MAM loss is defined as the cross-entropy between predicted probabilities and ground-truth masked tokens $\mathbf{Y}$ conditioned on the visual Action Tokens $\mathbf{S}$:
\begin{equation}
\mathcal{L}_{mam} = - \sum_{t=1}^{N-1} \log P(y_{t+1} \mid \mathbf{h}_{1:t}, \mathbf{S})
\end{equation}

where $N$ is the sequence length. This objective encourages the model to infer missing verbs and action sequences by attending to motion cues, thereby mitigating the ambiguity often found in noisy instructional transcripts.
\begin{figure}[!htb]
    \centering
    \includegraphics[width=0.85\linewidth]{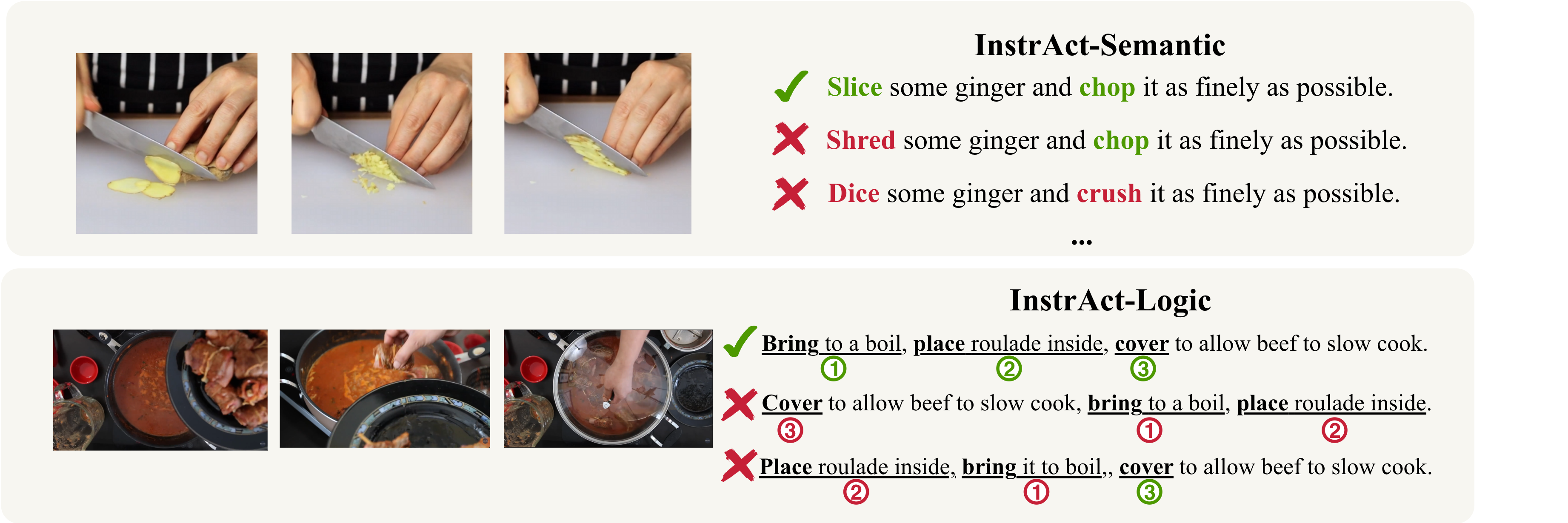}\\[0.2em]
    {\small (a) Example Multiple-Choice Questions (MCQ) for InstrAct-Semantic and InstrAct-Logic.}\\[0.8em] 
    
    \begin{minipage}[b]{0.50\textwidth}
        \centering
        \includegraphics[width=0.95\linewidth]{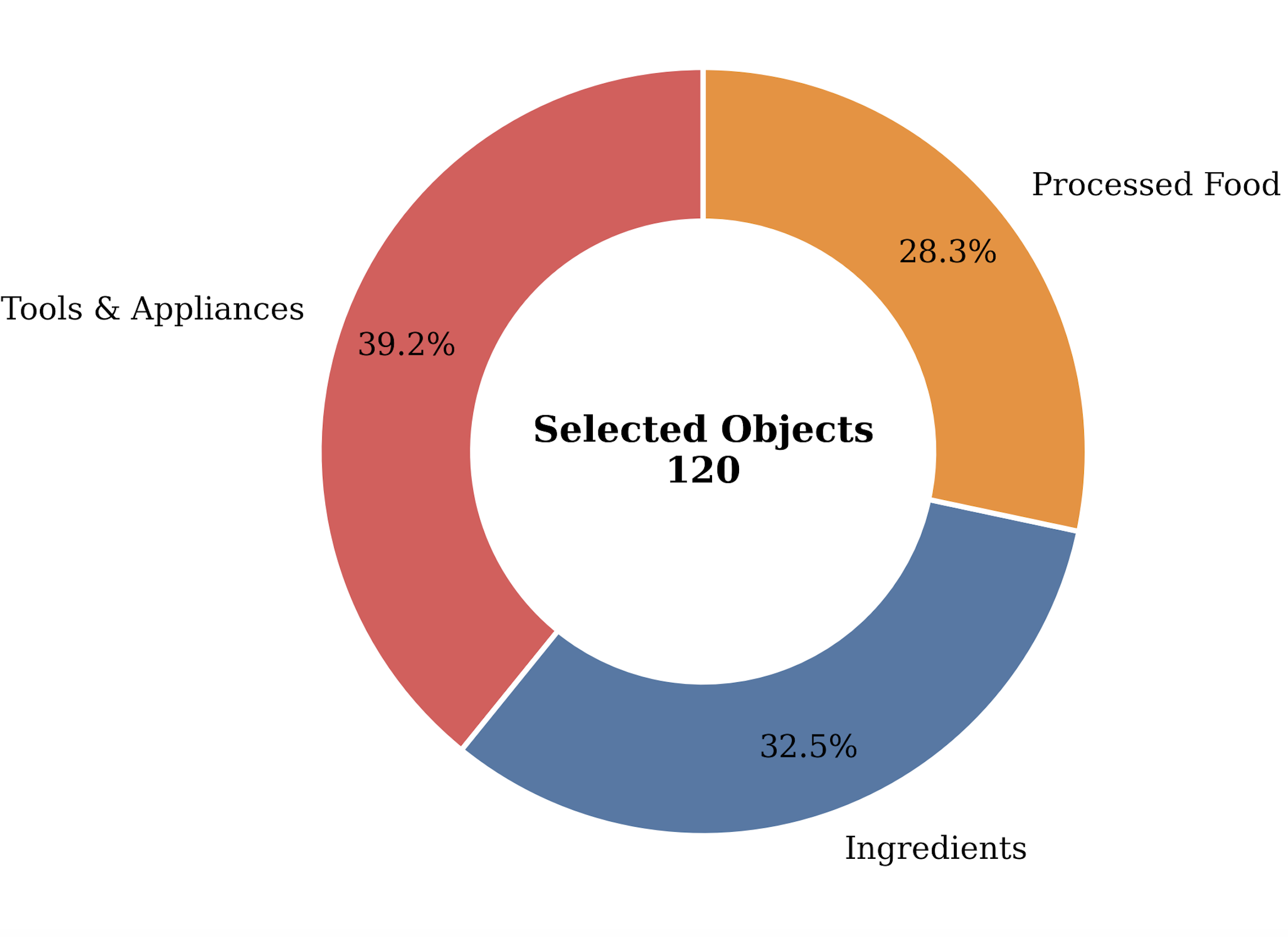}\\[0.2em]
        {\small (b) Category distribution of \textit{InstrAct-Dynamics}.}
    \end{minipage}
    \hfill 
    \begin{minipage}[b]{0.45\textwidth}
        \centering
        \includegraphics[width=0.95\linewidth]{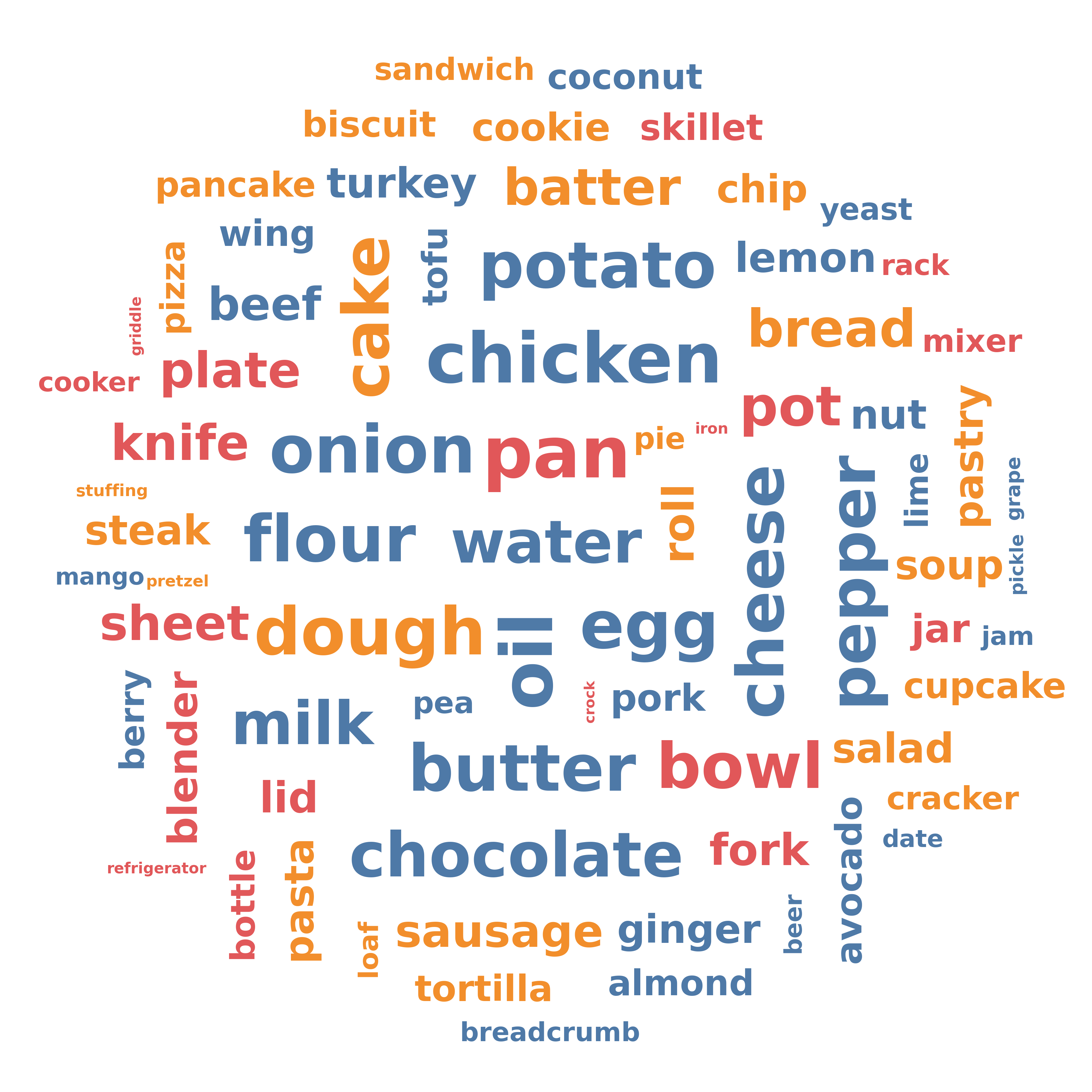}\\[0.2em]
        {\small (c) Word cloud visualization of the evaluated objects.}
    \end{minipage}

    \vspace{0.5em} 
    \caption{\textbf{Data Visualization.} Top: Illustrative examples from our semantic and logic benchmarks. Bottom: Statistical breakdown of the dynamics benchmark, showcasing the diverse object pool and category proportions.}
    \label{fig:overall_dataset_stats}
\end{figure}

\section{InstrAct Bench}
To evaluate whether video-language models truly understand instructional procedures rather than relying on static visual shortcuts, we introduce the \textbf{InstrAct Bench}, a benchmark constructed from the LLM-refined dataset (Sec.~\ref{sec:data}). The benchmark consists of three complementary tasks: InstrAct-Semantic for fine-grained verb discrimination, InstrAct-Logic for procedural reasoning, and InstrAct-Dynamics for action-centric cross-modal retrieval. The formulation and evaluation metrics are described below.

\subsection{InstrAct-Semantic}
\label{subsec:semantic}

InstrAct-Semantic diagnoses whether models rely on static visual shortcuts instead of motion cues. We formulate this task as a 10-choice multiple-choice question (MCQ) task with 2,000 video instances (Fig.~\ref{fig:overall_dataset_stats}, top). Given a video, the candidate answers consist of the ground-truth caption and nine action-altered hard negatives that preserve identical objects and context (Sec.~\ref{sec:data}). By controlling object semantics, this task forces models to distinguish fine-grained actions purely through motion dynamics (e.g., whisking eggs vs. cracking eggs). Performance is evaluated using Recall@1 (R@1), Recall@5 (R@5), and Mean Rank (MR).

\subsection{InstrAct-Logic}
\label{subsec:logic}

InstrAct-Logic evaluates whether models capture procedural dependencies in multi-step actions. This task is also formulated as a multiple-choice question (MCQ) task with 2,000 video instances. Each example contains the ground-truth caption and 2–3 order-altered negatives that permute the chronological order of actions while preserving the same content (Fig.~\ref{fig:overall_dataset_stats}, bottom). This design isolates temporal reasoning: models that rely on bag-of-actions representations may succeed on InstrAct-Semantic but fail here without modeling correct action order. Performance is evaluated using Accuracy (ACC).

\subsection{InstrAct-Dynamics}
\label{subsec:dynamics}

To complement synthetic evaluation with real-world data, we introduce InstrAct-Dynamics, a fine-grained cross-modal retrieval benchmark consisting of 16,326 video-text pairs grouped into 120 object-centric pools. Within each pool, all videos share the same primary object, forcing models to distinguish subtle procedural variations rather than relying on object recognition.

To capture diverse interaction types, the pools are categorized into Ingredients (raw state changes, e.g., cutting), Processed Food (complex transformations, e.g., baking), and Tools \& Appliances (functional manipulation). Retrieval is evaluated in both Video-to-Text and Text-to-Video settings within each pool. We report the clip-count-weighted average Recall@K (K = 1, 5, 10).

\section{Experiments}
\subsection{Experiment Setting}
We evaluate a diverse set of video-language models, including early contrastive approaches (e.g., MIL-NCE~\cite{miech19endtoend}) and recent Video Foundation Models (e.g., PerceptionLM~\cite{cho2025perceptionlm}).Baselines are categorized as In-Domain (ID) or Out-of-Domain (OOD) depending on whether their pretraining data includes cooking instructional videos (e.g., YouCook2). ID models are evaluated in a zero-shot setting, while OOD models are first fine-tuned on a filtered cooking subset before evaluation. Unless otherwise specified, InstrAction uses the ViCLIP-B branch of InternVideo as the default backbone.

Existing baselines are trained only with matched video–text pairs and do not utilize hard negatives (HNs). Directly evaluating our model using the same HN types seen during training would introduce an unfair advantage. To ensure a fair comparison, we adopt a cross-evaluation strategy. Specifically, for InstrAct-Semantic we train with Order-Swapped HNs, while for InstrAct-Logic we train with Verb-Altered HNs. For InstrAct-Dynamics, which contains only authentic video–text pairs without synthetic negatives, all HN types are used during training.

To further validate generalizability, we integrate InstrAction with four different VFM backbones and observe consistent improvements. Additional implementation details are provided in the Appendix.

\begin{table}[!htbp]
    \centering
    \footnotesize 
    \setlength{\tabcolsep}{4pt} 
    \caption{\textbf{Performance on InstrAct-Semantic and InstrAct-Logic.} Methods are grouped as In-domain (ID) and Out-of-domain (OOD). $\downarrow$ indicates lower is better.}
    \label{tab:instr_act_results_restructured}
    \resizebox{0.83\linewidth}{!}{
    \begin{tabular}{l ccc c c}
        \toprule
        & \multicolumn{3}{c}{\textbf{InstrAct-Semantic}} && \textbf{InstrAct-Logic} \\
        \cmidrule(lr){2-4} \cmidrule(lr){6-6}
        \textbf{Method} & \textbf{R@1 (\%)} & \textbf{R@5 (\%)} & \textbf{MR} ($\downarrow$) && \textbf{ACC (\%)} \\
        \midrule
        \rowcolor{gray!10} \multicolumn{6}{l}{\textit{ID Models (Zero-shot)}} \\
        MIL-NCE~\cite{miech19endtoend} & 7.2 & 49.4 & 6.0 && 5.1 \\
        UNiVL~\cite{Luo2020UniVL} & 13.2 & 60.3 & 5.0 && 33.6 \\
        VideoPrism~\cite{zhao2024videoprism} & 18.2 & 74.4 & 4.0 && 31.3 \\
        PerceptionLM~\cite{cho2025perceptionlm} & 16.5 & 63.2 & 4.0 && 31.4 \\
        ViCLIP~\cite{wang2023internvid} & 14.7 & 61.2 & 4.0 && 30.0 \\
        \midrule
        \rowcolor{gray!10} \multicolumn{6}{l}{\textit{OOD Models (FT on our Filtered Cooking Subset(175690 Clips))}} \\
        CLIP-ViP~\cite{xue2023clipvip} & 18.4 & 63.2 & 4.0 && 24.5 \\
        CLIP4Clip~\cite{luo2022clip4clip} & 14.8 & 58.0 & 5.0 && 23.7 \\
        \midrule
        \rowcolor{gray!10} \multicolumn{6}{l}{\textit{Our Methods}} \\
        InstrAction (Verb-Altered HNs) & - & - & - && \textbf{40.0} \\
        InstrAction (Order-Swapped HNs) & \textbf{28.1} & \textbf{78.0} & \textbf{3.0} && - \\
        \bottomrule
    \end{tabular}
    } 
\end{table}

\subsection{Main Results}
\paragraph{InstrAct-Semantic}
On the InstrAct-Semantic benchmark, VideoPrism~\cite{zhao2024videoprism} and PerceptionLM~\cite{cho2025perceptionlm} achieve the strongest performance among ID models, suggesting that large-scale pretraining combined with action-aware modeling improves sensitivity to motion dynamics. In contrast, MIL-NCE~\cite{miech19endtoend} performs poorly when confronted with our hard negatives, despite being pretrained on HowTo100M, indicating that simple video–text matching is insufficient for distinguishing fine-grained actions. Among OOD methods, CLIP-ViP slightly outperforms CLIP4Clip, reflecting modest gains from large-scale video–text pretraining. Nevertheless, our proposed method achieves the best performance across all metrics, demonstrating its effectiveness in disentangling actions from static object cues.

\paragraph{InstrAct-Logic}
On the InstrAct-Logic benchmark, most ID models (except MIL-NCE) achieve similar performance levels. This suggests that scaling training data or adopting specialized architectures alone is insufficient under the prevalent holistic video–text matching paradigm. MIL-NCE performs substantially worse on this task, indicating that simple video–text matching models struggle with order-sensitive action sequences. Since the order-altered negatives share nearly identical vocabularies with the ground-truth captions, distinguishing them requires modeling fine-grained temporal dependencies. OOD methods exhibit a similar trend even after domain-specific fine-tuning, further highlighting the limitations of standard matching objectives for modeling procedural logic. In contrast, our method achieves a clear performance improvement, demonstrating its effectiveness in capturing sequential action dependencies.

\begin{table}[!htbp]
\centering
\scriptsize
\renewcommand{\arraystretch}{1.0}

\caption{\textbf{Overall performance on the InstrAct-Dynamics benchmark.}
Weighted average Recall@K across 120 objects (16,326 clips).}
\label{tab:dynamics_results_overall}

\resizebox{0.83\linewidth}{!}{
\begin{tabular}{lccc ccc ccc}
\toprule
\multirow{2}{*}{\textbf{Method}}
& \multicolumn{3}{c}{\textbf{R@1}}
& \multicolumn{3}{c}{\textbf{R@5}}
& \multicolumn{3}{c}{\textbf{R@10}} \\

\cmidrule(lr){2-4}
\cmidrule(lr){5-7}
\cmidrule(lr){8-10}

& T2V & V2T & Mean
& T2V & V2T & Mean
& T2V & V2T & Mean \\

\midrule
MIL-NCE~\cite{miech19endtoend}      
& 12.63 & 15.04 & 13.84
& 30.77 & 34.42 & 32.59
& 42.10 & 46.69 & 44.40 \\

UniVL~\cite{Luo2020UniVL}        
& 13.77 & 15.91 & 14.84
& 32.67 & 36.80 & 34.73
& 43.89 & 49.14 & 46.51 \\

PerceptionLM~\cite{cho2025perceptionlm}
& 16.33 & 23.10 & 19.72
& 32.11 & 42.05 & 37.08
& 41.15 & 51.48 & 46.32 \\

VideoPrism~\cite{zhao2024videoprism}   
& 27.23 & 34.13 & 30.68
& 48.49 & 59.08 & 53.79
& 58.36 & \textbf{70.00} & 64.18 \\

\midrule
\textbf{Ours}
& \textbf{33.86} & \textbf{35.19} & \textbf{34.53}
& \textbf{54.73} & \textbf{59.52} & \textbf{57.13}
& \textbf{63.13} & 69.71 & \textbf{66.42} \\

\bottomrule
\end{tabular}
}
\end{table}

\begin{figure}[!htbp]
  \centering
  \includegraphics[width=0.8\linewidth]{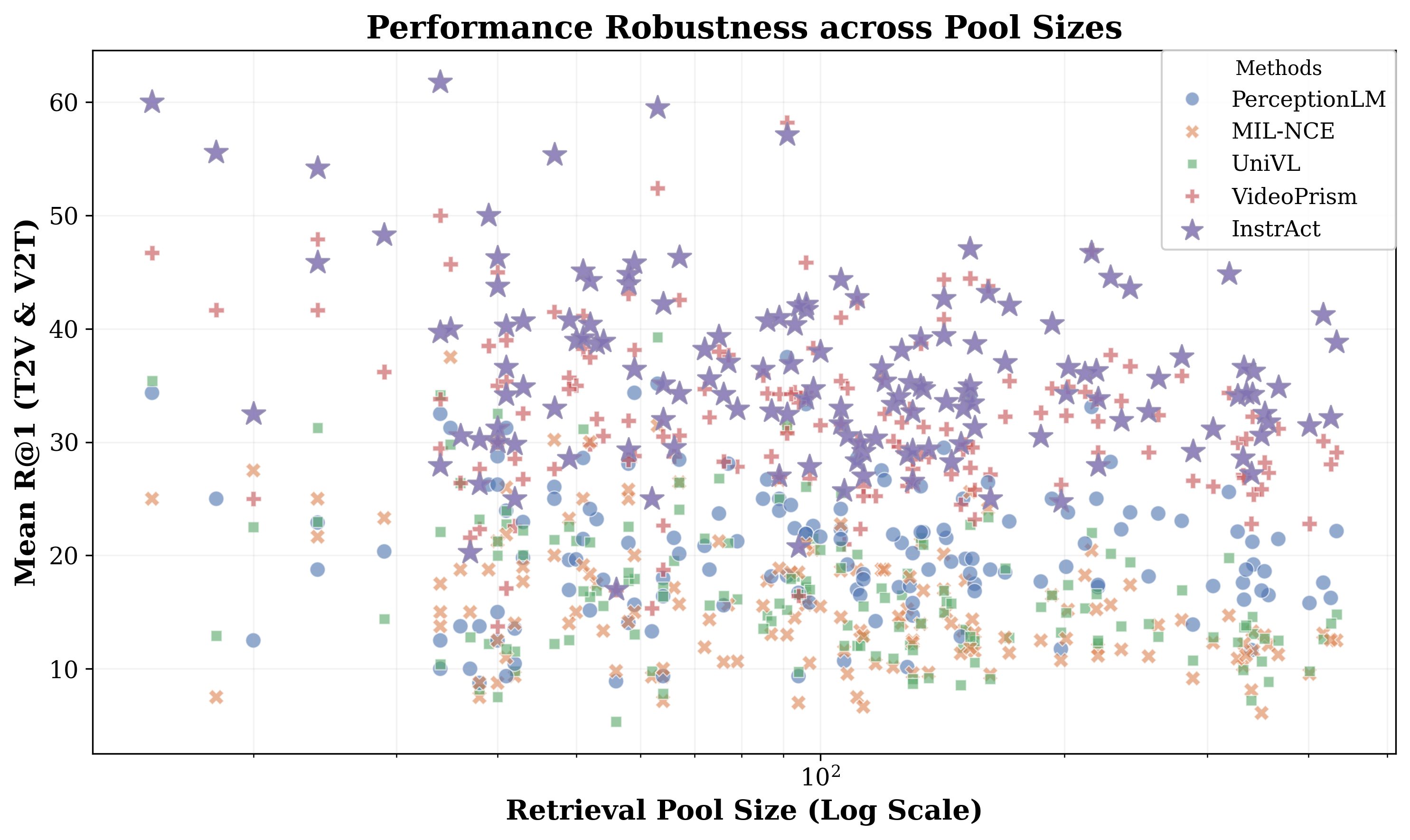}
  \caption{Qualitative results of text-to-video retrieval on InstrAct-Dynamics pools.}
  \label{fig:retrieval_vis}
\end{figure}

\paragraph{InstrAct-Dynamics}
\label{subsec:dynamics-exp}
Table~\ref{tab:dynamics_results_overall} reports performance on authentic videos (detailed breakdowns in the Appendix). Our model achieves the best results on nearly all metrics, improving mean R@1 from 30.68 to 34.53 and mean R@5 from 53.79 to 57.13 over the strongest baseline (VideoPrism). Because each pool contains videos with the same primary object, this benchmark requires models to distinguish subtle procedural variations rather than rely on static object cues. VideoPrism also performs strongly, especially in V2T retrieval, consistent with its motion-aware design. Category-level analysis in the Appendix further shows that our model gains the most on Tools \& Appliances scenarios, where fine-grained functional manipulation is critical.

Figure~\ref{fig:retrieval_vis} further analyzes retrieval robustness under different pool sizes. As the number of candidates increases, most baselines degrade noticeably due to the growing number of in-domain distractors. In contrast, our model maintains consistently higher accuracy across pool scales, demonstrating stronger robustness in fine-grained action discrimination.

\begin{table}[!htbp]
\centering
\footnotesize
\caption{\textbf{Generalization across different Video Foundation Models.} We evaluate the performance gains of applying solely our data-driven augmentation (+ HNs only) versus deploying our complete InstrAction framework which incorporates both data- and model-driven approaches (+ InstrAct) across four distinct backbones.}
\label{tab:generalization}

\resizebox{0.70\linewidth}{!}{
\begin{tabular}{llcccc}
\toprule
\multirow{2}{*}{\textbf{Backbone}} & \multirow{2}{*}{\textbf{Method}} & \multicolumn{3}{c}{InstrAct-Semantic} & InstrAct-Logic \\
\cmidrule(lr){3-5} \cmidrule(lr){6-6}
& & R@1 (\%) & R@5 (\%) & MR ($\downarrow$) & ACC (\%) \\
\midrule
          & Backbone      & 7.2  & 49.4 & 6.0 & 5.1  \\
MIL-NCE~\cite{miech19endtoend}   & + HNs only     & 32.4 & 82.2 & 3.0 & 6.1  \\
          & + InstrAction & 32.9 & 83.3 & 2.0 & 4.7  \\
\midrule
          & Backbone      & 18.4 & 63.2 & 4.0 & 24.5 \\
CLIP-ViP~\cite{xue2023clipvip}  & + HNs only     & 38.5 & 86.7 & 2.0 & 30.2 \\
          & + InstrAction & 42.3 & 88.7 & 2.0 & 37.1 \\
\midrule
          & Backbone      & 14.8 & 58.0 & 5.0 & 23.7 \\
CLIP4Clip~\cite{luo2022clip4clip} & + HNs only     & 44.7 & 90.4 & 2.0 & 35.6 \\
          & + InstrAction & 43.3 & 88.8 & 2.0 & 40.2 \\
\midrule
          & Backbone       & 14.7 & 61.2 & 4.0 & 30.0 \\
ViCLIP~\cite{wang2022internvideo}    & + HNs only     & 40.7 & 89.4 & 2.0 & 37.2 \\
          & + InstrAction & 45.1 & 90.8 & 2.0 & 44.7 \\
\bottomrule
\end{tabular}
}
\end{table}

\paragraph{Generalization Ability}
Table~\ref{tab:generalization} demonstrates the generalizability of InstrAction across four VFM backbones. Applying only our data-driven augmentation (+HNs) consistently improves the base models; for example, CLIP-ViP improves from 18.4 to 38.5 and CLIP4Clip from 14.8 to 44.7 in R@1 on InstrAct-Semantic. Incorporating the full framework (+InstrAction) further brings improvements for most backbones, especially on the more challenging InstrAct-Logic task, indicating that both the data-driven augmentation and the action-centric modeling contribute to stronger action understanding. However, MIL-NCE shows negligible gains and even slight drops on InstrAct-Logic, which aligns with our earlier observations on InstrAct-Dynamics: its relatively simple text encoder struggles to capture order-sensitive dependencies when captions share nearly identical vocabularies.

\subsection{Ablation Study}

\begin{figure}[!htbp]
    \centering
    \includegraphics[width=0.95\linewidth]{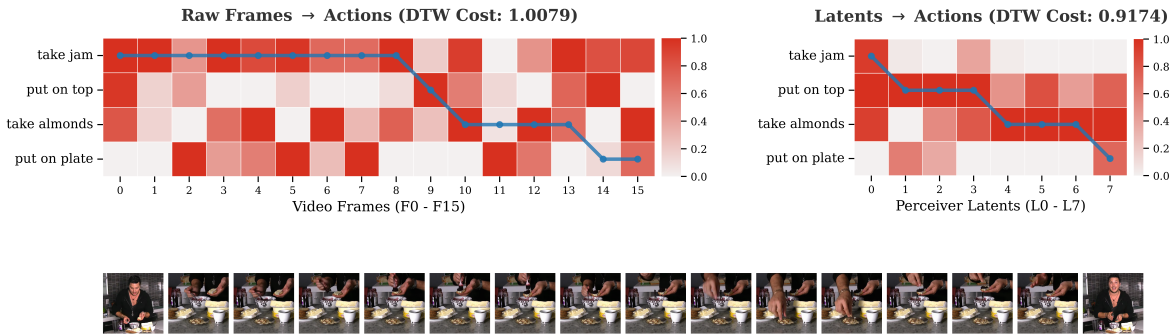}
    \caption{Qualitative comparison of DTW alignment heatmaps. Left subplots show Baseline (Raw Frames $\rightarrow$ Actions), and Right subplots show (Latents $\rightarrow$ Actions).}
    \label{fig:perceiver_heatmap}
\end{figure}

\paragraph{Analysis on Action Perceiver}
\begin{wrapfigure}{r}{0.45\textwidth}
    \vspace{-22pt}
    \centering
    \setlength{\tabcolsep}{4pt}
    \renewcommand{\arraystretch}{1.05}

    \resizebox{0.9\linewidth}{!}{
    \begin{tabular}{lc}
    \toprule
    \textbf{Visual Representation} & \textbf{Cost} $\downarrow$ \\
    \midrule
    Raw Embeddings & 0.97 \\
    Action Latents & \textbf{0.91} \\
    \bottomrule
    \end{tabular}
    }

    \caption{Quantitative comparison of alignment cost (lower is better).}
    \label{tab:dtw_cost}
    \vspace{-7pt}
\end{wrapfigure}
To validate the Action Perceiver, we compare it with a baseline using raw frame embeddings on a subset of videos containing more than three sequential verb phrases. Alignment quality is measured using the normalized Dynamic Time Warping (DTW) cost:
\begin{equation}
\mathcal{C}_{DTW}=\frac{1}{L}\sum_{(i,j)\in P^*}(1-S_{i,j})
\end{equation}
where $S\in\mathbb{R}^{T\times V}$ denotes the similarity matrix between $T$ visual tokens and $V$ verb phrases, and $P^*$ is the alignment path. As shown in Table~\ref{tab:dtw_cost}, Perceiver latents yield a lower alignment cost than raw embeddings, indicating reduced temporal noise and better alignment with sequential actions.

\begin{table}[!htb]
    \centering
    \caption{\textbf{Ablation Study on InstrAct Bench.} We evaluate the impact of Hard Negatives (HN), and 2 proposed constrains. $\downarrow$ indicates lower is better.}
    \label{tab:ablation_study_clean}
    \renewcommand{\arraystretch}{1.2} 
    \setlength{\tabcolsep}{8pt} 
    \begin{tabular}{@{} ccc ccc c @{}} 
        \toprule
        \multicolumn{3}{c}{\textbf{Components}} & \multicolumn{3}{c}{\textbf{InstrAct-Semantic}} & \textbf{Logic} \\
        \cmidrule(lr){1-3} \cmidrule(lr){4-6} \cmidrule(l){7-7}
        \textbf{HN} & \textbf{Action Perceiver} & \textbf{Align.} & \textbf{R@1} & \textbf{R@5} & \textbf{MR} ($\downarrow$) & \textbf{ACC} \\
        \midrule
        \checkmark & --         & --        & 40.7 & 89.4 & 2.0 & 37.2 \\
        \checkmark & \checkmark & MAM       & 43.7 & 90.0 & 2.0 & 39.4 \\
        \checkmark & \checkmark & DTW-Align & 38.7 & 86.5 & 2.0 & 40.2 \\
        \midrule
        \checkmark & \checkmark & Both      & \textbf{45.1} & \textbf{90.8} & \textbf{2.0} & \textbf{44.7} \\
        \bottomrule
    \end{tabular}
\end{table}

Qualitatively (Figure~\ref{fig:perceiver_heatmap}), the baseline exhibits scattered activations caused by static object bias, often correlating common objects such as ``jam'' with all frames. In contrast, the Action Perceiver produces a cleaner staircase-like diagonal pattern that isolates sequential steps (e.g., ``put on''). These results suggest that the Action Perceiver yields temporally cleaner representations; we thus keep it fixed and ablate the training objectives.

\paragraph{Ablation on Training Objectives}
Table~\ref{tab:ablation_study_clean} analyzes the effect of the proposed training objectives. Starting from the baseline trained with hard negatives only, adding MAM improves InstrAct-Semantic (R@1: 40.7$\rightarrow$43.7) by encouraging fine-grained cross-modal reconstruction. In contrast, DTW-Align mainly benefits InstrAct-Logic (ACC: 37.2$\rightarrow$40.2) by enforcing order-sensitive temporal alignment, although it slightly degrades Semantic performance where local verb discrimination is more critical. When both objectives are jointly optimized, the model achieves the best overall results (R@1: 45.1, ACC: 44.7), indicating that temporal grounding and masked semantic reconstruction provide complementary supervision for robust action understanding.

\section{Conclusion}
We address the object shortcut bias in instructional videos by shifting from static matching to action-centric modeling. By combining a refined HowTo100M subset with our Action Perceiver and DTW-based alignment, we successfully decouple fine-grained action semantics from redundant background context. Our framework and the new InstrAct Bench demonstrate that while data scale is helpful, explicit architectural priors and high-quality action supervision are essential for understanding how procedural activities structurally unfold.

%
\bibliographystyle{plain}
\bibliography{main}
\end{document}